\pdfoutput=1

\documentclass[a4paper, 10pt, conference]{ieeeconf}      

\usepackage[utf8]{inputenc}
\usepackage[T1]{fontenc}
\usepackage{hyperref}
\hypersetup{colorlinks,allcolors=black}
\usepackage{booktabs}
\usepackage{graphicx}
\usepackage{subfig}

\title{\LARGE \bf
G-Rank: Unsupervised Continuous Learn-to-Rank for Edge Devices in a P2P Network\\
\vspace{3mm}

\small \--- MSc. Thesis \---

}

\author{ \parbox{3 in}{\centering Andrew Gold\\
        Delft University of Technology\\
        Delft, The Netherlands\\
        {\tt\small a.w.r.gold@student.tudelft.nl}}
    \and
    \parbox{3 in}{\centering Dr.ir. J.A. Pouwelse\\
        Delft University of Technology\\
        Delft, The Netherlands\\
        {\tt\small j.a.pouwelse@tudelft.nl}}
}

\begin{document}

\maketitle
\thispagestyle{plain}
\pagestyle{plain}

\begin{abstract}

Ranking algorithms in traditional search engines are powered by enormous training data sets that are meticulously engineered and curated by a centralized entity.
Decentralized peer-to-peer (p2p) networks such as torrenting applications and Web3 protocols deliberately eschew centralized databases and computational architectures when designing services and features. 
As such, robust search-and-rank algorithms designed for such domains must be engineered specifically for decentralized networks, and must be lightweight enough to operate on consumer-grade personal devices such as a smartphone or laptop computer.
We introduce G-Rank, an unsupervised ranking algorithm designed exclusively for decentralized networks.
We demonstrate that accurate, relevant ranking results can be achieved in fully decentralized networks without any centralized data aggregation, feature engineering, or model training. 
Furthermore, we show that such results are obtainable with minimal data preprocessing and computational overhead, and can still return highly relevant results even when a user's device is disconnected from the network.
G-Rank is highly modular in design, is not limited to categorical data, and can be implemented in a variety of domains with minimal modification.
The results herein show that unsupervised ranking models designed for decentralized p2p networks are not only viable, but worthy of further research.
\newline
\textit{Author's note: the experiments performed herein are open-source and can be found on GitHub\footnote{\hspace{1mm}https://www.github.com/awrgold/G-Rank}}.

\end{abstract}


\section{INTRODUCTION}


The problem of relevance ranking in information retrieval problems has been well-studied for decades, solutions for which have enabled users to query vast swathes of information on the World Wide Web and retrieve highly relevant results within milliseconds.
Nascent search-and-rank techniques for web search culminated with PageRank in 1998 \cite{pagerank}, directly leading to Google's ascendant dominance in the web search domain. 
All such algorithms, however, depend upon ever-growing databases of mapped relations between various information sources and topics, requiring enormous computational power to deliver lightning-fast results directly to a user's device.
Therefore, these algorithms all depend upon highly centralized information architectures with thousands of skilled attendants dedicated to maintaining and improving system capabilities.
In such a paradigm, the risk of impropriety such as misdirection and fraud is high due to the enormous financial incentives for being ranked higher in search results.


As such, typical ranking algorithms are wholly unsuited for deployment in decentralized information architectures such as peer-to-peer (p2p) file sharing networks (e.g. BitTorrent) and various Web3 applications.
These networks are largely comprised of individual users where the maximum computational and storage capacity available to any search-and-rank algorithm is that of an individual's desktop computer or mobile device.
The success of many nascent applications built atop decentralized networks therefore depends upon the efficacy of novel search-and-rank schemes designed specifically for edge devices in these domains.
These algorithms must have a zero-server architecture, be lightweight enough to run on a cheap smartphone, and yet be robust enough to return highly relevant results to each individual user.

Furthermore, these algorithms must adhere to the ethos of these decentralized networks, which often emphasize user privacy and information security foremost among its tenets. 
Any ranking algorithm built in such a domain must therefore be able to function effectively utilizing data immediately available to a user of a p2p application, the majority of which is often the user's own data. 
That is not to say that a ranking algorithm cannot be improved via the sharing of information between participants in such networks, but rather that the algorithm must be entirely self-sufficient and self-contained without any meaningful expectation of obtaining new information outside of the local device.
As first proposed in 2013 by Ormándi et al. \cite{gossip learning}, the concept of utilizing message-passing as a means to build a cohesive machine learning model in a distributed setting became a novel instrument in respecting user privacy by emphasizing local-first computational paradigms.

The concept of local-first software is not new \cite{local-first}, and privacy-preserving machine learning schemes such as encrypted machine learning \cite{encrypted ml 1}\cite{encrypted ml 2} and federated machine learning \cite{federated ml 0}\cite{federated ml 1}\cite{federated ml 2}\cite{federated ml 3} already exist, yet the problems of security, storage, and overhead persist.
Unfortunately, most of these machine learning models are supervised which handicaps developers by requiring large amounts of high-quality training data to achieve meaningful results.
Furthermore, many unsupervised ranking models that show promising results \cite{unsupervised ranking 1}\cite{unsupervised ranking 2}\cite{unsupervised ranking 3}\cite{unsupervised ranking 4} are designed exclusively for centralized systems.
As such, any decentralized algorithm or model that can quickly and sufficiently retrieve and rank search results without the need for model training or human supervision would allow for p2p networks of any size to deliver meaningful search capabilities in a more trustless fashion.
Therefore, truly decentralized unsupervised ranking system sits at the forefront of p2p and Web3 communications development.

The rapid growth of p2p file-sharing networks around the turn of the new millennium led to a boom in research for search algorithms designed explicitly for such networks \cite{cubit}\cite{hyperspaces}\cite{semantic p2p}\cite{decentralized search}\cite{decentralized search 2}\cite{decentralized pagerank}\cite{collaborative search}\cite{decentralized unsupervised 1}. 
Many such algorithms attempted to recreate the efficacy of well-known existing search and rank algorithms such as PageRank, yet the number of publications plateaued and began to decline around 2012. 
The explosive growth of blockchain and Web3 technologies has influenced a new generation of developers designing for a more decentralized web experience.
Decentralized search and rank algorithms that do not depend upon any centralized entity to function properly, are domain-independent, and can sufficiently replicate the performance of more centralized solutions are still nascent.
We demonstrate that a simple, lightweight, and effective ranking algorithm can be deployed to p2p applications while achieving respectable results.

We introduce the unsupervised ranking algorithm G-Rank designed explicitly for ranking search results in an internet-deployed p2p torrent-based music streaming platform.
The goal of this first validation experiment is to demonstrate the "correctness" of an unsupervised learn-to-rank (LTR) model in the context of a distributed p2p file sharing network. 
This model requires no training data to function, is capable of returning relevant results to users within the first few queries, and is not constrained by any dependence upon large datasets.
G-Rank is demonstrably capable of ranking results in line with their global popularity, even though the model itself is unaware of the best possible ranking for any given query term.
G-Rank will quickly approach the optimal global ranking for all peers in the network, even if a user does not perform any queries themselves; as a network utilizing G-Rank grows in usership, new users will see highly relevant results even with their first query.

The rest of this paper is as follows. 
\hyperref[sec:problem]{Section 2} expounds upon the problem of relevance ranking, namely supervised versus unsupervised methods.
\hyperref[sec:approach]{Section 3} details the implementation of the G-Rank algorithm, describing the clicklog structure and gossip-based information dissemination mechanism necessary for its functioning, as well as the experimentation and evaluation of the model.
\hyperref[sec:simulations]{Section 4} describes a number of experimental simulations of p2p network participants under a variety of scenarios, including the results of each experiment.
\hyperref[sec:conclusion]{Section 5} concludes that our algorithm is capable of deployment, and provides suggestions for future work.


\section{PROBLEM DESCRIPTION}
\label{sec:problem}

Security within the domain of decentralized machine learning remains an unsolved problem.
There exist numerous additional constraints in decentralized networks that traditional machine learning models need not be concerned with.
Trustless, anonymous networks are rife with malevolent usership, and the task of identity verification in such networks also remains unsolved.
Adding an additional layer of complexity, many p2p networks are built upon open-source software, affording any would-be adversary direct insight into potential attack vectors.
As such, machine learning models engineered for public p2p networks require exceptional attention to detail across all facets of their design.
These constraints disqualify any supervised models from the outset as they violate the trustless nature of p2p networks. 
Either the engineers of such supervised models must be trusted to train and validate the model, or the network participants must provide training data themselves, thereby introducing a critical vulnerability.
Creating a LTR search engine for a p2p domain that requires no training yet can converge towards an optimal ranking as if an error rate is being minimized in a supervised model would constitute a major development in p2p applications.

Learn-to-rank is a well-known and thoroughly-studied problem with myriad solutions achieving excellent results, yet many of the most well-known ranking algorithms are designed around centralized data aggregators and supervised training methods.
Past research into ranking search results within p2p networks are almost exclusively supervised methods \cite{LTR partially labeled}\cite{LTR gradient descent}\cite{CF for LTR}\cite{OG LTR}, which besides the traditional pitfalls mentioned above also constrain the ranking problem into an optimization problem.
Furthermore, such supervised methods lack inherent "memory" such that they cannot retain new information as they observe it; as such, they require large training sets and trusted providers of training data.
Compiling relevant datasets and appropriate labels requires considerable effort, which historically has been performed manually by humans and is infeasible for exceptionally large datasets.
Automated labeling methods such as semi-supervised learning can speed up this process, but these methods have the drawback of imparting their own inherent bias into the constructed dataset \cite{semi-supervised 1}\cite{semi-supervised 2}.
Therefore, the difficulty of labeling data in a manual or semi-supervised manner grows faster relative to the increase in size of data.


Other solutions treat ranking as a recommendation prediction problem, where results are sorted by the predicted score \cite{decentralized CF}\cite{CF movie ranking}\cite{advances in CF ranking}\cite{CF for LTR}.
Framing the ranking problem as a recommendation prediction problem also depends heavily on the manner in which users "score" items that they are recommended.
Depending on the application, the manner in which scores are calculated heavily influences the behavior of the recommender. 
In the domain of e-commerce, an item purchased by a user may be assigned a higher score than an item said user has viewed multiple times but not purchased, even if the user feels that the viewed item is more relevant to them. 
Meanwhile, a music recommender may assign a higher score to a song that appears in multiple playlists of a specific user yet has fewer overall streaming plays than a song that does not appear in any playlist yet contains a significant number of streaming plays for that same user.
As such, any scoring system must be thoughtfully designed for the specific recommendation algorithm and its domain.

With regards to distributed machine learning, federated machine learning has several drawbacks in this domain as well.
Federated models are often less accurate due to their relative inability to capture the variance in the overall data throughout the network, as each model is iteratively fitted to a small subset of data.
Federated learning techniques, as presented in \cite{federated ml 0}\cite{federated ml 1}\cite{federated ml 2}\cite{federated ml 3}, utilizes message passing to disseminate model parameters during training.
This parameter-passing mechanism is often considered sufficient enough to obfuscate local data - affording some degree of user privacy - though such methods are insufficient to prevent determined adversaries from recreating input data \cite{FML recreation blog}.
That being said, any such supervised methods still face the issue of requiring training datasets which limits the scope of potential research due to inadequate training data availability and the infeasibility of synthesizing such datasets oneself.
As such, unsupervised ranking algorithms that can approach the performance of supervised ranking methods may be better suited towards p2p domains, where a significant portion of software is open-source and user privacy is often given higher priority than for traditional web services.
Significantly reduced overhead in algorithm implementation and maintenance, therefore, is of major benefit to p2p applications.

Machine learning models deployed in distributed or decentralized settings are vulnerable to several specific attack vectors, namely sybil and spam attacks, which can undermine model accuracy and efficacy, e.g. via "model poisoning attacks" \cite{poisoning fml 1}\cite{poisining fml 2}. 
Such attacks are inherently difficult to thwart in any decentralized network setting.
As shown in \cite{pagerank sybil}, even PageRank is not immune to sybil attacks and therefore also requires considerable adaptation to trustless p2p environments.
Sybil attacks on federated machine learning models present critical vulnerabilities, and solutions such as those mentioned in \cite{fools gold} depend upon assumptions that are unobtainable in live p2p networks.
Meanwhile, spam attacks are often broader in scope yet still pose significant risk to machine learning models whose efficacy depend upon the veracity of the data they are fed.

These threats are well-understood and a variety of methods to thwart such attacks exist \cite{fools gold}\cite{mitigating sybil 1}\cite{mitigating sybil 2}, however many of these solutions are based on supervised learning and therefore suffer from the same issues mentioned previously, or require the aggregation of network traffic through centralized "coordinators," eroding the trustlessness of p2p networks.
As such, unsupervised machine learning models that are robust enough to function in the midst of spam or sybil attacks are critical to the expansion of search, ranking, and recommendation models for the decentralized web.

\begin{figure}
    \centering
    \includegraphics[scale=0.5]{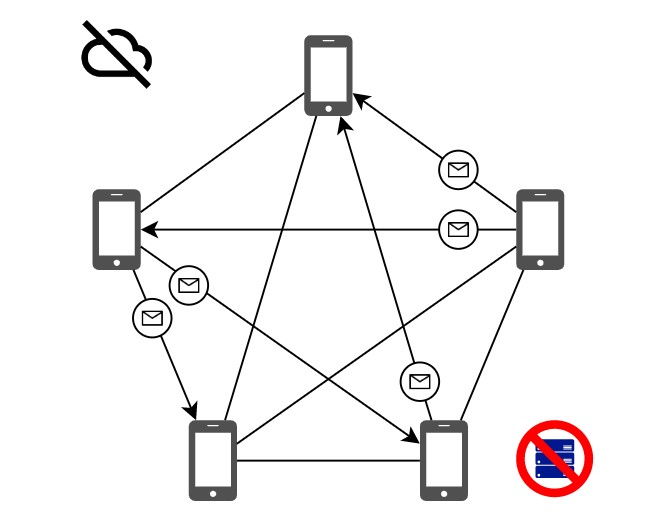}
    \caption{Decentralized p2p networks are zero-server architectures where often the only mechanism of information dissemination is via message-passing, infusing an additional constraint into the machine learning architecture.}
    \label{fig:my_label}
\end{figure}

\section{ARCHITECTURE OF G-RANK}
\label{sec:approach}

Our G-Rank algorithm is a first humble step towards a first decentralised search engine. 
We focus on the domain of music and video search specifically. 
Our p2p architecture assumes each user operates their own node and searches for BitTorrent-based Creative Commons licensed music.
This music application allows users (A.K.A. "nodes" when referring to network architecture, or "peers" when referring to other users in the network) to query other peers for the contents of their library and download files to their device.
The clicklog is the central data structure within our architecture.
It contains the user query and supporting info.
Whenever a user issues a query, the user device appends the query and its associated results to a clicklog that is stored locally on the device.
At any point, each peer can request an update from another peer containing its local clicklog, disseminating clicklog data with other peers in the network via a gossip protocol (See \hyperref[sec:gossip]{Section 3B}).
When a user receives a gossip message containing updated clicklog information, the device appends the new information to the local clicklog to be used by the ranking model in future queries.

The unsupervised method detailed herein focuses on ranking query search results relative to one another, i.e. pairwise comparison across all potential results.
Due to the fact that each node in the network contains only a small subset of total possible search results, it is highly unlikely that any one node attain perfect ranking results without the dissemination of local clicklog information to other nodes in the network.
Such a mechanism - be it via gossip, broadcasting search history, or a centralized information aggregation scheme - directly and heavily influences the behavior of the unsupervised ranking model.
The continuous updating of data accessible to G-Rank is an example of continuous learning \cite{continuous learning}, where the model requires no re-training as each new data point becomes available.
Instead, as each gossip message is received G-Rank considers this new information in real time, affording it the ability to continuously adapt to an ever-changing environment with zero human intervention.
Therefore, the ranking model's dependence upon the clicklog dissemination scheme is closely investigated alongside the actual performance of the ranking model, where two distinct gossip schemes are considered alongside ranking model parameters and functionality.

\subsection{Unsupervised Ranking Model}
\label{sec:model}

When a user searches for a query term, the ultimate goal is to provide the most accurate list of results ranked by relevance to the query term as well as to the user. 
First, the model checks the local clicklog for previous instances of a query term, and if this term has never been queried before it then searches for matches of this term in the metadata of local files, including the title, artist, and genre tags.
The model does not consider misspellings/typos, although methods such as those mentioned in \cite{cubit} are highly effective at correcting for typos in information retrieval (IR) schemes and could potentially be integrated with G-Rank.
If the query term has been seen before, it returns the most popular results for this query weighted by the similarity of search and click behavior of other users who have also issued similar or identical queries (as described in \hyperref[sec:similarity]{Section 3D}).
In order to avoid plateauing performance, G-Rank incorporates a degree of statistical noise by swapping two randomly-selected items in the list of results for 50\% of the queries.

Due to the fact that the search mechanism considers only the clicklog and item metadata, it is extremely unlikely that a item should erroneously become popularly associated with a query term that has no direct match with any of the item's metadata.
The only situation in which this could arise is if a query term has never been seen before nor is contained in any accessible metadata. 
Should this happen, the search engine returns a list of popular items that have appeared recently in the user's local clicklog.
However, because peers have the option to share their local clicklogs with other peers upon request, it is entirely plausible that a node or subset of the network could be unaware of newly added items with matching metadata at the time of the query.
If this were to occur, a user could click on a recommended item that contains no matching metadata to the search query and then gossip their clicklog history to nearby nodes, who then also perform a search for the same term and click on the same result.
Such an occurrence would then erroneously lead to a term-item pairing for which the associated item actually contains no matching metadata, which could then propagate throughout the network.

In order to avoid this situation, search results that contain matching metadata are always ranked above items that have term-item matches in the clicklog yet contain no matching metadata. 
The justification for such is that should users wish to find a specific item, they are ostensibly aware of the title, artist, album, or some other trait that would be found within the item's metadata such that they need not rely entirely upon the search history of a specific term in order to find said item.
A positive side-effect of this restriction is that it also diminishes the effect of adversarial users "query-bombing" the network to negatively influence the performance of the ranking model.

\subsection{Clicklog Structure}
\label{sec:clicklog}

\begin{figure}
    \centering
    \includegraphics[scale=0.3]{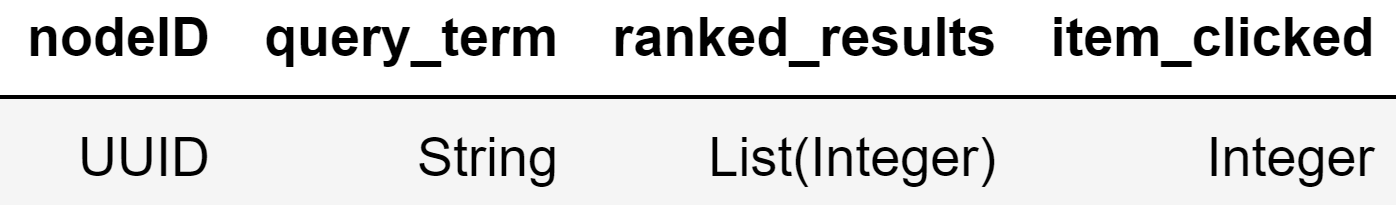}
    \caption{The primary attributes of the clicklog data structure. Each entry contains a unique identifier of the node performing the query, the query term, the ranked results (in descending order) for the query, and the item clicked upon.}
    \label{fig:clicklog}
\end{figure}

Each node in the network contains a \hyperref[fig:clicklog]{clicklog} that stores the following primary attributes as a row entry: the node's unique ID, the query term, the query results in descending order, and the item the user clicked on.
Additional clicklog attributes include the title of the item clicked upon, the tag metadata associated with that item, and a unique key associated with the query term consisting of the concatenation of the node's unique ID and the local query number.
These additional attributes are used primarily during the evaluation of the simulations, though G-Rank does consider tag metadata during ranking if the querying node's clicklog does not reflect any direct query term matches.
When a query is performed, the results are stored in local memory until a user clicks upon a result, after which the clicklog entry is created and appended locally.
Over time, each node becomes increasingly aware of the click behavior of other nodes in the network without necessarily gleaning insight into the local libraries of said nodes.
As such, the dissemination of clicklog data enables the unsupervised model to learn from the behavior of other users without revealing personally identifiable information.

\subsection{Gossiping Clicklogs}
\label{sec:gossip}

Gossip-based protocols allow for dissemination of information throughout a p2p network with varying degrees of efficiency.
Regarding G-Rank, it is understood that traditional unsolicited gossip propagation schemes present a clear and present attack vector for adversaries to undermine the model's performance.
Therefore, G-Rank depends upon \textit{solicited gossip} for clicklog dissemination.
At any time, any node can send a \textit{request} message to one or more nodes it is aware of, also known as a \textit{pull} gossip scheme.
The recipient of a \textit{request} message may reply with a \textit{response} message containing some or all of its local clicklog, which may contain clicklog entries from other nodes that the recipient has received via issuing its own \textit{request} messages.
In our experiments, nodes cannot refuse update requests and only send \textit{request} messages to a single node.

The design of the gossip protocol that propagates clicklog information directly affects the performance of the ranking model, and therefore needs to be deliberately designed such that clicklog information is adequately disseminated without congesting the network. 
In order to determine exactly how the gossip parameters affect the model, specific evaluation metrics need to be determined. 
For example, should a node receive $|K|=10$ results for a specific query, it is important to determine how many of these results are in the "optimal" ranking, i.e. for each result $k_i \in K$ the distance between the local rank $L(k_i)$ in the above query versus the global average rank $G(k_i)$ across all participants in the network for that query term.
In this situation, an item with an "optimal" ranking has a distance of $G(k_i) - L(k_i) = 0$ for any specific query.

In distributed and decentralized networks it is well-understood that obtaining a global "snapshot" of the current network state becomes intractible as the network grows large.
Well-known algorithms such as Chandy-Lamport \cite{chandy lamport} are still imperfect as they fail to capture incipient changes to the network state deriving from messages that are currently underway during the time of the snapshot, such that by the time the algorithm terminates the state of the network may have already changed.
As such, determining a global truth for a p2p network can only be easily performed in a contained simulation environment in which a global observer aggregates all changes to the network's state.
Therefore, it must be understood that any comparison against a "global" optimum in our experiments comes with the caveat that in a live network the global optimum may not be feasibly observable.

\subsection{Node Discovery and Similarity Clustering}
\label{sec:similarity}

The primary unsupervised method in G-Rank is based on a fuzzy non-parametric semantic self-clustering of nodes based upon a pairwise similarity score as described below. 
When a gossip response is received by node $n_i$ it appends the new data to its existing clicklog and updates its local list of known nodes in the network.
Such is the mechanism of node discovery in the network: via the receipt of clicklog data from other nodes in the network.

After receiving a clicklog update, $n_i$ searches the incoming data for previously unseen unique node IDs.
These unseen node IDs are added to a local list of known nodes, which are then sorted in descending order by a modified Jaccard similarity score between their queries and the results they each click upon.
The similarity score $S$ between a pair of nodes $n_i$ and $n_j$ is calculated as follows:

\begin{itemize}
    \item Find the cardinality of the intersection of the top $K$ query terms $T^{K}(Q)$ between $n_i$ and $n_j$, denoted as $\kappa_{t}$.
    \item Find the square of the cardinality of the intersection of clicked results for all query terms $C_i(Q)$ and $C_j(Q)$ between $n_i$ and $n_j$, denoted as $\kappa_{m}$.
    \item The sum $\kappa_{t} + (\kappa_{m})^{2}$ is divided by the cardinality of the the union of clicked results for all query terms $C_i(Q)$ and $C_j(Q)$ between $n_i$ and $n_j$, denoted as $\kappa_{u}$.

\end{itemize}

That is, 

$$ S_{i}(n_j) = \frac{\kappa_{t} + (\kappa_{m})^{2}}{\kappa_{u}} $$

The list of scores $S_i(N)$ is normalized by dividing by $\verb|max|(S_i)$ resulting in a similarity score between $0.0$ and $1.0$, where $1.0$ indicates that two nodes have clicked on the exact same item for every single matching query.
Therefore, the similarity score is a weighted ratio of identical query-click tuples to the overall number of queries shared between two nodes.
As such, every node maintains a list of nodes it has become aware of via the clicklog, and determines its similarity to other nodes based on past click behavior. 
This similarity is then used to weight the results of future queries based on the click behavior of other users, such that users are more likely to see results other similar users have clicked on for similar query terms.
By including $(\kappa_{m})^{2}$ in the similarity score, we account for divergent click behavior such that node similarity scores follow an exponential gradient.
If the click behavior of node $n_i$ diverges from that of $n_j$ over time, $S_{i}(n_j)$ will more rapidly decrease than otherwise, allowing for more expedient "re-clustering."

In order to isolate highly divergent click behavior, we introduce the isolation constant $F$ to the user similarity score.
When $F=0$, only the clicklogs of adjacent nodes with $S_i(n_j) > 0$ are considered when ranking results.
When $F=1$, a node considers with equal weight the clicklog entries of all nodes it has received gossip from when ranking query results.
As such, this isolation parameter allows for nodes to discount the clicklogs of other nodes if these nodes have query and click behavior that does not match its own at least once.
Similarity weighting is calculated by taking the dot product between the aforementioned similarity scores for each node and sorted results based on the overall number of clicks found in each node's local clicklog.
The resulting ranking $R$ provided to querying node $n_i$ for query $Q$ is therefore calculated as:

$$R_{i}(Q) = (\forall k \in K_Q), \hspace{2mm}  \sum_{j=0}^{N} (C_k \cdot (S_{i}(n_j) + F))$$

where $S_i(n_j)$ indicates the similarity score for each node pair $(n_i, n_j) \in N$, and $C_k$ indicates the number of clicks associated with item $k \in K_Q$ where $K_Q$ is the unsorted set of results for query $Q$.
The resulting items are sorted in descending order by their associated scores.
As such, each potential query result is assigned a score based on the number of clicks found in each node's clicklog, weighted by the similarity of each node to the node performing the query.
Therefore, the dissemination of clicklog data not only informs other nodes of the popularity of items, it also allows for nodes to cluster themselves based on an easy-to-compute metric, further allowing for personalization of results.

A potential drawback of introducing such a similarity metric into the ranked results is that it introduces a possible attack vector for adversaries to influence the results of future queries throughout the network, e.g. via spam or sybil attacks.
Spam attacks become less viable as the number of legitimate users grows larger, while more targeted attacks may be thwarted by the user similarity scheme itself.
An adversary attempting to undermine the ranking algorithm by intentionally selecting irrelevant results for specific queries would find themselves increasingly isolated from other users performing legitimate queries, as their behavior over time would continue to deviate from that of other users.
Sophisticated adversaries would then need to mimic legitimate behavior for a large portion of their queries in order to remain relevant to other users without ostracizing themselves.

\section{Dataset and Experiment Setup}
\label{sec:setup}

Our dataset consists of actual music releases and associated metadata. 
Our experimental setup is tailored to minimize the work to deploy G-Rank for decentralised search of BitTorrent audio and video content.

\subsection{Dataset}
\label{sec:dataset}

The dataset utilized in this experiment was compiled from a series of 256 actual music releases by real artists via the PandaCD record label\footnote{ \hspace{1mm} https://pandacd.io/}, all of which were released under the Creative Commons license.
Entries may be singles, albums, EPs (extended-play releases), and LPs (limited-play releases).
Every entry consists of three attributes: \textit{Title}, \textit{Artist}, and \textit{Album}, as well as a number of associated \textit{Tags} as metadata, which describe the release in terms of genre.
These tags have been compiled into a corpus of potential query terms, and every query term in this experiment consists of exactly one tag, of which there are a total of 39 unique values.

\subsection{Experiment Setup}
\label{sec:experimentation}

For all experiments we conducted an evaluation round every 100 queries, where a number of performance metrics are gathered (see \hyperref[sec:metrics]{Section 5F)}.
In addition to the regular performance evaluation, these evaluation rounds act as progress markers at discrete intervals in the simulation, which are discussed in \hyperref[sec:analysis]{Section 5G}.
Each experiment, including the baseline, was conducted twice: once with similarity weighted isolation constant $F=0$ and again with $F=1$, demonstrating the effect that cluster isolation (see \hyperref[sec:similarity]{Section 3D}) has on G-Rank's performance.
Unless stated otherwise, simulation parameters are as follows:

\begin{itemize}
    \item All gossip targets are drawn exclusively from each node's local clicklog data.
    
    \item All nodes keep track of gossip \textit{progress} such that previously-shared clicklog contents are omitted from new gossip requests.
    
    \item When a new node joins the network, it is bootstrapped by a randomly-selected node who shares with it a randomly-sampled subset of its own clicklog.
    Via this bootstrap mechanism, each adversarial node becomes aware of a handful of other nodes in the network to which it can gossip during its attack phase.
    \item There are exactly 10 malicious nodes in each adversarial experiment (with the exception of the Epic Sybil Attack), which are bootstrapped as stated above at simulation time step $t=2500$, exactly 25\% through the simulation. 
\end{itemize}

Across all experiments, the simulated network consists of 100 nodes, all of which begin with a limited number of library items. 
The simulation is initialized as follows. 
For each node $n_i \in N, i = \{0,...,99\}$ in the network, $n_i$ is initialized by selecting at uniform random $10\%$ of the items from the music dataset to add to its local library (approximately 26 songs per node).
Next, a series of initial queries are performed.
For each of the 39 possible query terms, each node performs a search for said query term and chooses at random one item from its library with a tag matching the query term and appends this entry to its local clicklog.
Should a node's library not contain any items with tags matching the query term, it selects at random a single item from its local library, thereby introducing a small degree of noise into the clicklog.
At this point, no ranking or click modeling is utilized for selection, and the clicklog of node $n_i$ contains exactly 39 items.
Then, node $n_i$ gossips a random sample of 10\% of its local clicklog to node $n_{i-1}$ such that each node contains no more than 44 clicklog entries; 39 belonging to itself, and up to an additional five items that it receives via gossip from another peer.

This method of initialization affords each peer in the network an even number of clicklog items to utilize during a query, but an uneven distribution of network knowledge across each node such that nodes with higher IDs are more likely to be aware of a higher number of peers at the outset of the simulation.
After every node has been initialized, the simulation begins and nodes are chosen uniformly at random alongside a random query term from the corpus to perform a query-term search.
The results of the search are ranked as detailed in \hyperref[sec:model]{Section 2A}, and an item to be clicked upon is chosen based on the \hyperref[sec:click model]{aforementioned click model}.
The search and click results are then appended to this node's local clicklog.
Thereafter, this node then performs a gossip round by requesting a gossip update from a randomly-selected peer node it is aware of (except in the case of the \textit{Push vs. Pull} experiments, see \hyperref[sec:pushpull]{Section 5D}).

There are two popular schemes for initiating gossip in p2p networks: time-based and probabilistic.
In time-based schemes, a node gossips every $t$ time units, whereas in most probabilistic schemes any given node has a probability $p$ per time unit to gossip some information to a subset of other nodes, such that after $t$ time units there is a $$\Pr(X=t)=(1-p)^{t-1} \cdot p$$ probability that a node will have gossiped.
To clearly illuminate the effect of adversaries on G-Rank's performance, our experiment implements a hybrid gossip approach such that at every simulation tick $t$ a random node is receiving at least one update from another node it is aware of (see \hyperref[sec:gossip]{Section 3C}).
As such, a node is guaranteed to receive gossip post-query yet still is chosen probabilistically such that the above geometric probability distribution holds, given that a node has probability $p = \frac{1}{|N|}$ of performing a query-then-gossip operation at an arbitrary time step $t$.
By utilizing such a gossip mechanism we ensure that clicklog information is propagated regularly throughout each simulation.

\subsection{Click Modeling}
\label{sec:click model}

Modeling realistic user-clicking behavior is essential to the development of ranking algorithms.
Not all user clicks may be on relevant items in a list, and as such it can be expected that a certain degree of noise exists in user click data.
Extrapolating such noise into a simulation therefore requires careful consideration.
Without anticipating and modeling a certain degree of noise, a ranking model's query results may erroneously converge towards irrelevant items.
Anticipating and modeling noise in user click behavior has been investigated \cite{noise click modeling}, however for this experiment users select the highest-ranking item in most queries, except when multiple results with equal relevance scores were shown to the user.
In this case, the result with the lowest item ID is chosen as a tiebreaker.

\section{Adversarial Simulations and Performance Analysis}
\label{sec:simulations}

We simulate several adversarial conditions alongside a baseline simulation with no adversaries.
Each adversarial simulation is intended to isolate and investigate the effects of specific adversarial and anti-social behavior on G-Rank's performance.
Each simulation's results is compared to the baseline global performance of G-Rank, as the global optimal rankings are negatively affected by such attacks.
As such, each scenario's impact on G-Rank's ability to converge towards a true global optimality without adversarial interference is investigated with the aid of the metrics described in \hyperref[sec:metrics]{Section 4B.}
In every adversarial simulation, the network is bootstrapped without any adversarial presence at first.
At time step $t=2500$, all attackers are bootstrapped into the network as described in \hyperref[sec:experimentation]{Section 4B}, where they lie in wait until time step $t=5000$ to begin their attack.
At this time, they perform their attack as described in each section below.
Post-bootstrap, these adversarial nodes may receive request messages from benign nodes, even if they have not yet performed their attack.

\subsection{Baseline Experiment}
\label{sec:baseline}

Initially we conduct a baseline validation experiment to demonstrate the sensitivity of the node discovery process within distributed machine learning. 
Realistic simulations lack any centrality and thus have no central coordinator to discover other nodes. 
Our design integrates node discovery via the clicklog itself using a unique node identifier.
Thus a single clicklog message provides both overlay network information for gossip dissemination, as well as the underlying data upon which the unsupervised model relies. 
This baseline experiment entails no adversarial interference, demonstrating how individual nodes adjust their rankings over time as they receive gossip throughout the simulation.
All other experiments build upon this validation simulation's setup for comparison purposes.

\subsection{Targeted Sybil Attack}
\label{sec:targeted}
The first adversarial simulation is performed to demonstrate how a sophisticated adversary could undermine a p2p network utilizing G-Rank by forcing irrelevant results towards the top of query results.
The adversaries execute a basic sybil attack where 10 new sybil nodes are bootstrapped into the network as described above, lying in wait until the predetermined attack time.
At the time of attack, each sybil attacker chooses a single specific term to perform 100 queries with, each time clicking the bottom-most item in the list of ranked results.
After the series of queries are complete, the attackers then lie in wait until they receive a gossip request from another peer in order to disseminate their malicious clicklog entries.
This attack artificially inflates the relevance of otherwise low-ranked results to a specific query term, undermining the veracity of the rankings other nodes are shown.
By repeatedly choosing the lowest-ranked item in the list, the attacker attempts to undermine G-Rank's ability to determine the most popular item associated with the query term.
The purpose of this experiment is to examine the effect deliberately misleading clicklog entries has on G-Ranks ability to converge towards optimality.

\subsection{Clicklog Inflation Attack}
\label{sec:inflation}
The purpose of the second adversarial simulation is to examine G-Rank's ability to re-converge towards optimal rankings after a sudden, significant growth in the number of clicklog entries that the model considers when ranking content.
This simulation differs from the \textit{Targeted Sybil Attack} in two key ways: the number of queries each adversary performs $1000$ queries instead of $100$, and each adversary chooses results purely at random.
By performing a significant number of queries before gossiping, the adversary attempts to undermine G-Rank by injecting a significant statistical noise into each node's clicklog.
The propagation of random clicklog noise throughout the network is investigated against the \textit{Targeted Sybil Attack} mentioned previously.

\subsection{Epic Sybil Attack}
\label{sec:persistent}
The purpose of this experiment is to examine the performance of G-Rank in the face of a significant number of adversaries.
The third adversarial simulation is nearly identical to the first, except that the number of attackers equals 75\% of the entire network participants.
Each node is bootstrapped in the same manner as the \textit{Targeted Sybil Attack}, lying in wait until the predetermined attack time.
At the time of attack, each node performs $100$ queries, again choosing the lowest-ranked item in the results for each query. 
They then wait until a gossip request message is received.
The inclusion of a network super-majority of sybil nodes is intended to investigate G-Rank's ability to improve rankings over time in the face of severe adversarial conditions.

\begin{figure*}
\centering
\setkeys{Gin}{width=0.45\linewidth}
\subfloat{\includegraphics{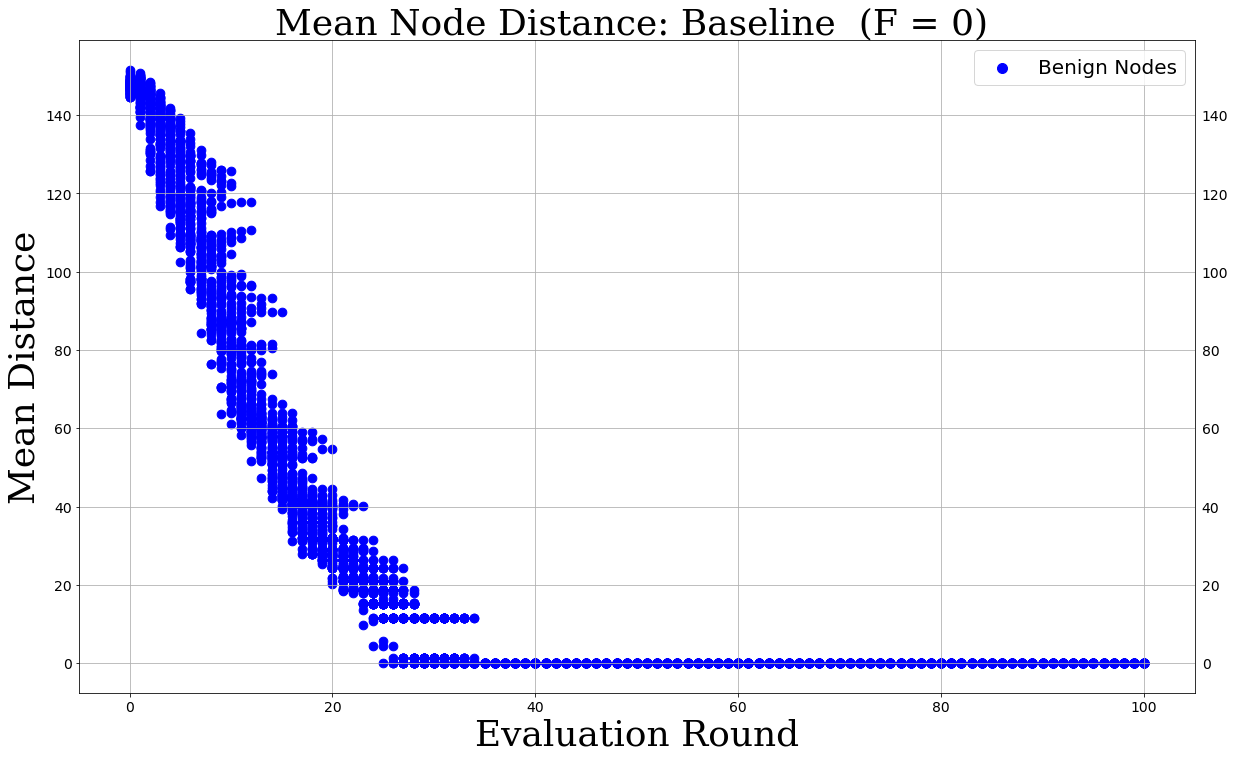}} 
\hfill
\subfloat{\includegraphics{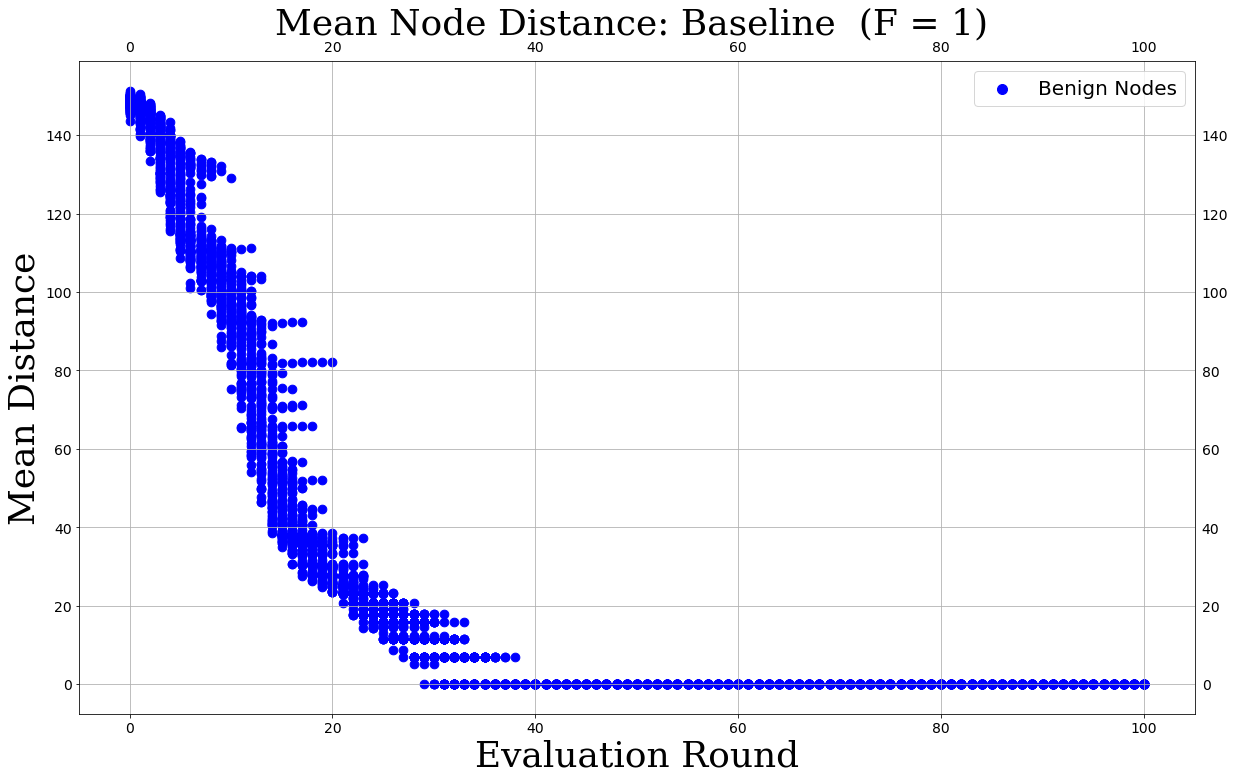}} 
\hfill

\caption{Scatter plots depicting each node's average distance to optimal rankings for each term, for each baseline simulation.}
\label{fig:baseline distance}
\end{figure*}

\subsection{Push vs. Pull Experiment}
\label{sec:pushpull}

Within this fourth experiment we introduce a number of malicious nodes which conduct a \textit{Targeted Sybil Attack} under a modified gossip scheme.
This experiment shows the dramatic impact of \textit{Push} versus \textit{Pull} gossip.
This experiment proves that is is vital for security that malicious nodes can not easily insert their polluting content with honest peers, \textit{i.e.} a push architecture.
With a pull architecture, peers are more autonomous and decide individually the speed of incoming information, if they trust another peer, or may randomly sample from discovered peers. 
Malicious nodes in this experiment send an unsolicited clicklog gossip messages to up to two peers\footnote{\hspace{1mm} In the rare circumstance that an adversary is only aware of one other peer, it only gossips to that peer.}, whereas benign nodes push to no more than one peer.
As Internet bandwidth is cheap, this simple experiment shows a first line of defence against clicklog spam without the need for significantly modifying G-Rank's core functionality.
With the pull architecture utilized in the original \textit{Targeted Sybil} attack, there exists only one recipient of a malicious node's gossip.
In this experiment, all nodes must accept incoming gossip messages.
By comparing the push gossip scheme to the original pull-based scheme, we illuminate the difference in G-Rank's convergence rate between two different gossip mechanisms without altering G-Rank's core functionality.
The purpose of this experiment is to highlight the effect various gossip and information dissemination schemes have upon G-Rank's efficacy.

\subsection{Evaluation Metrics}
\label{sec:metrics}

For each simulation we utilized a number of metrics to evaluate G-Rank's ranking performance over time, its tenacity when facing adversarial conditions, and the network capacity overhead over time.
The primary performance metric utilized is a positional edit distance metric where we compare the sum of index distances between each unique element in $R_i(Q)$ and $R_g(Q)$, where $R_g(Q)$ indicates the globally optimal ranking for query $Q$.
$R_g(Q)$ is computed simply by ranking the most popular items by their respective number of clicks associated with a specific query term across all nodes.
This metric allows us to determine how far each item is from its most optimal position at any given point in time, giving us the ability to determine how G-Rank performs for any given node for a specific query term.

We also consider the rate of G-Rank's convergence towards the global optimal averaged across all nodes and possible query terms over time.
The rate of change in this distance metric affords us insight into G-Ranks behavior over time, particularly during the adversarial simulations, such that we can better understand how G-Rank's long-term performance is affected by transient adversarial events.
For each possible query term we also measure the number of results containing the most popular result in the top position in order to demonstrate the roughly even distribution of performance, regardless of the frequency a specific query term is issued.

In terms of space and storage metrics, we also measure the average clicklog size across all nodes over time, as gossip occurs consistently yet as time goes on the number of duplicate clicklog items being shared likely continues to grow.
To better understand G-Rank's dependence upon gossip, we monitor the rate of growth in gossip message size (in bytes) as individual clicklogs grow large - an important metric considering the potential variation in each node's processing power and storage space.
However, we do not consider any time-based computational overhead metrics as these are highly dependent upon numerous factors, including the programming language in which G-Rank is implemented as well as each individual device's computational power.


\begin{figure*}
\centering
\setkeys{Gin}{width=0.3\linewidth}

\subfloat{\includegraphics{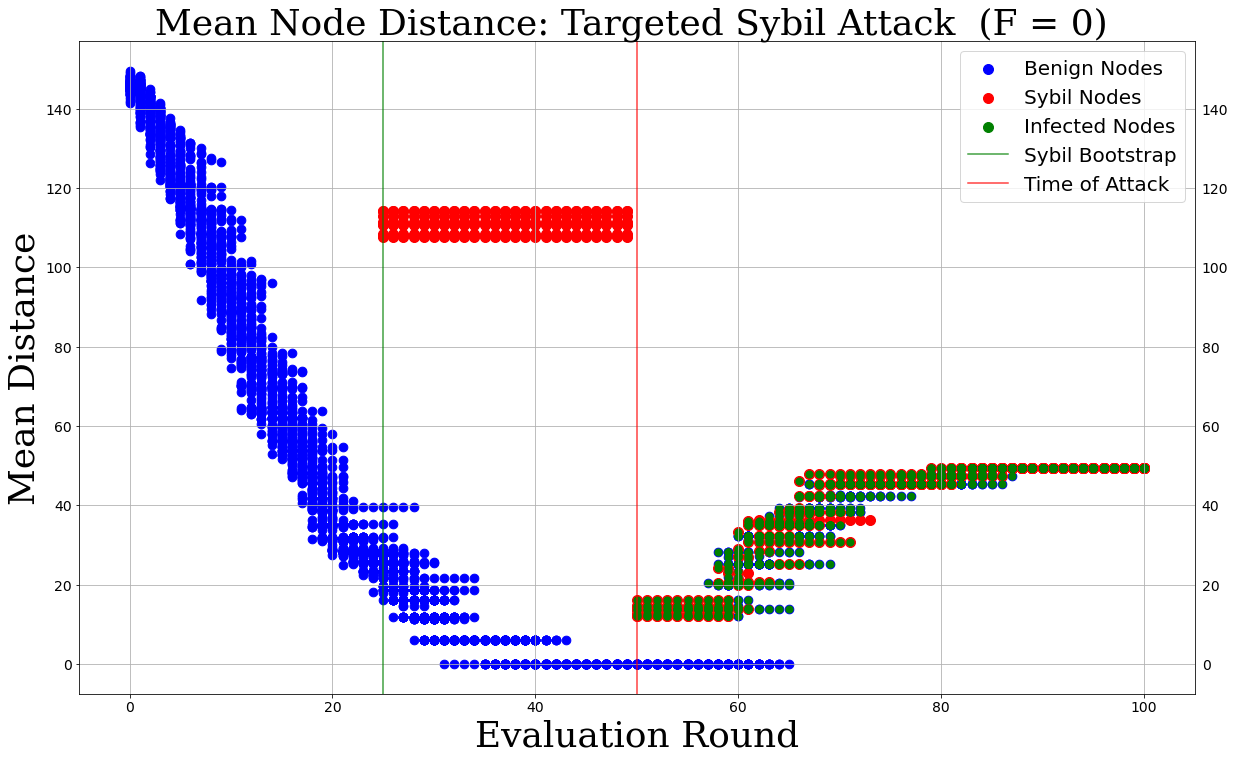}} 
\hfill
\subfloat{\includegraphics{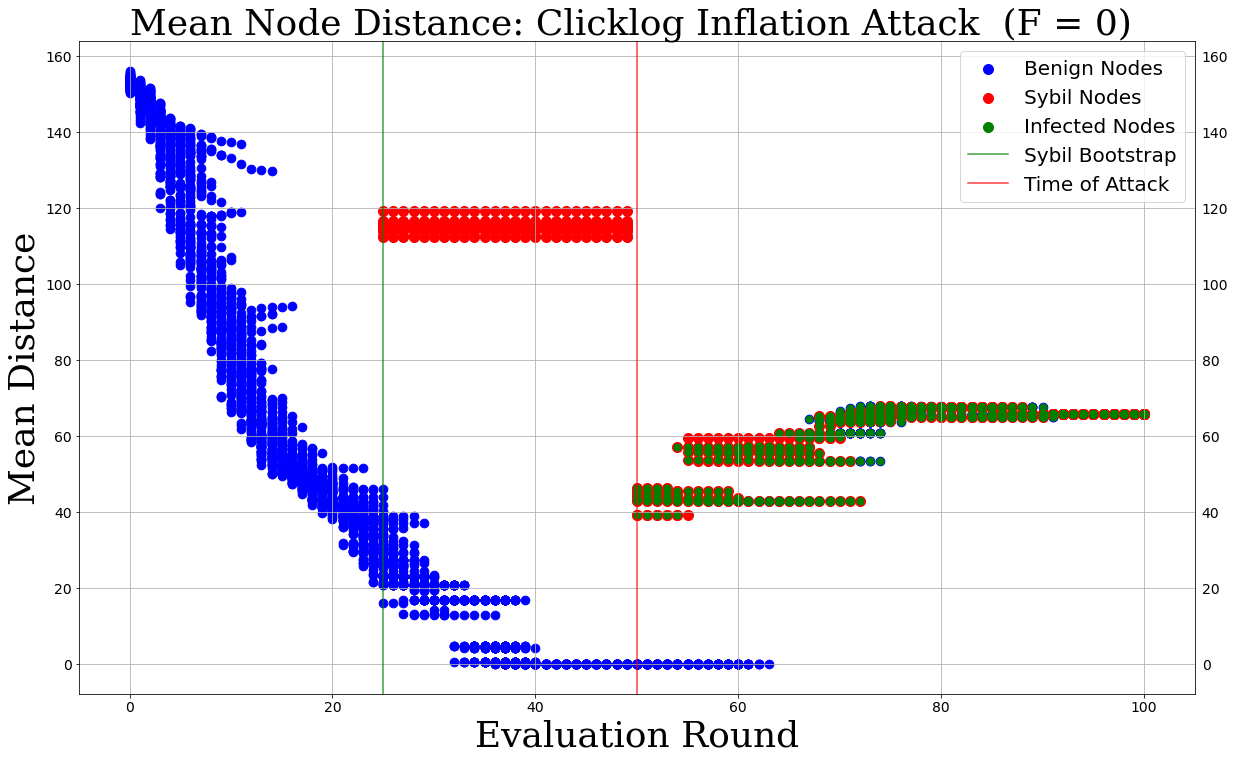}} 
\hfill
\subfloat{\includegraphics{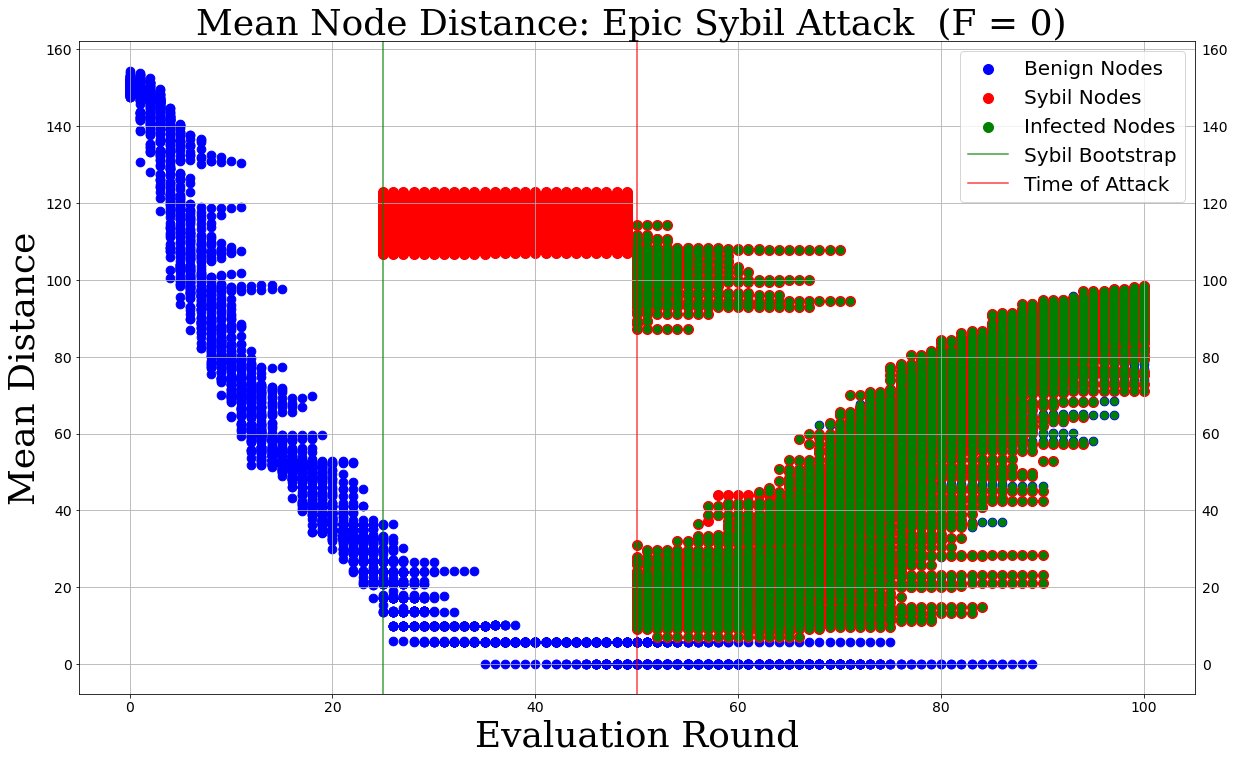}} 
\hfill

\vspace{5mm}

\subfloat{\includegraphics{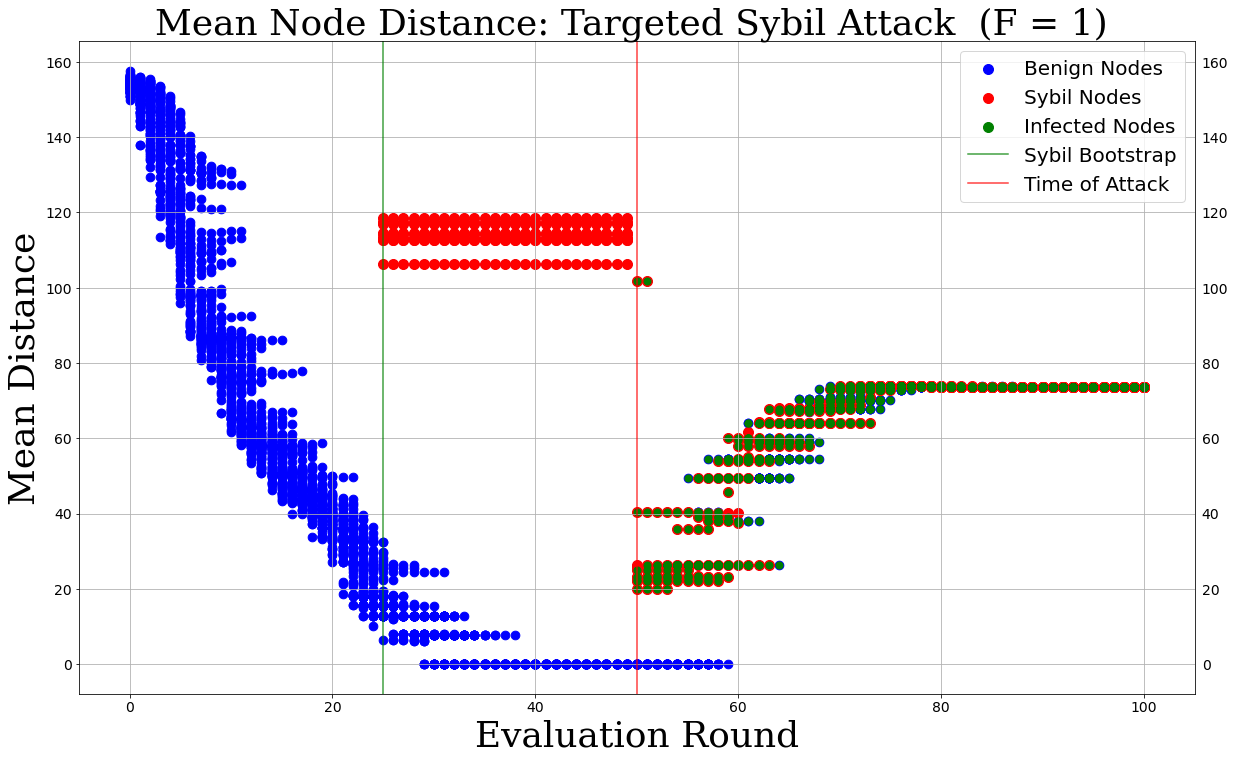}} 
\hfill
\subfloat{\includegraphics{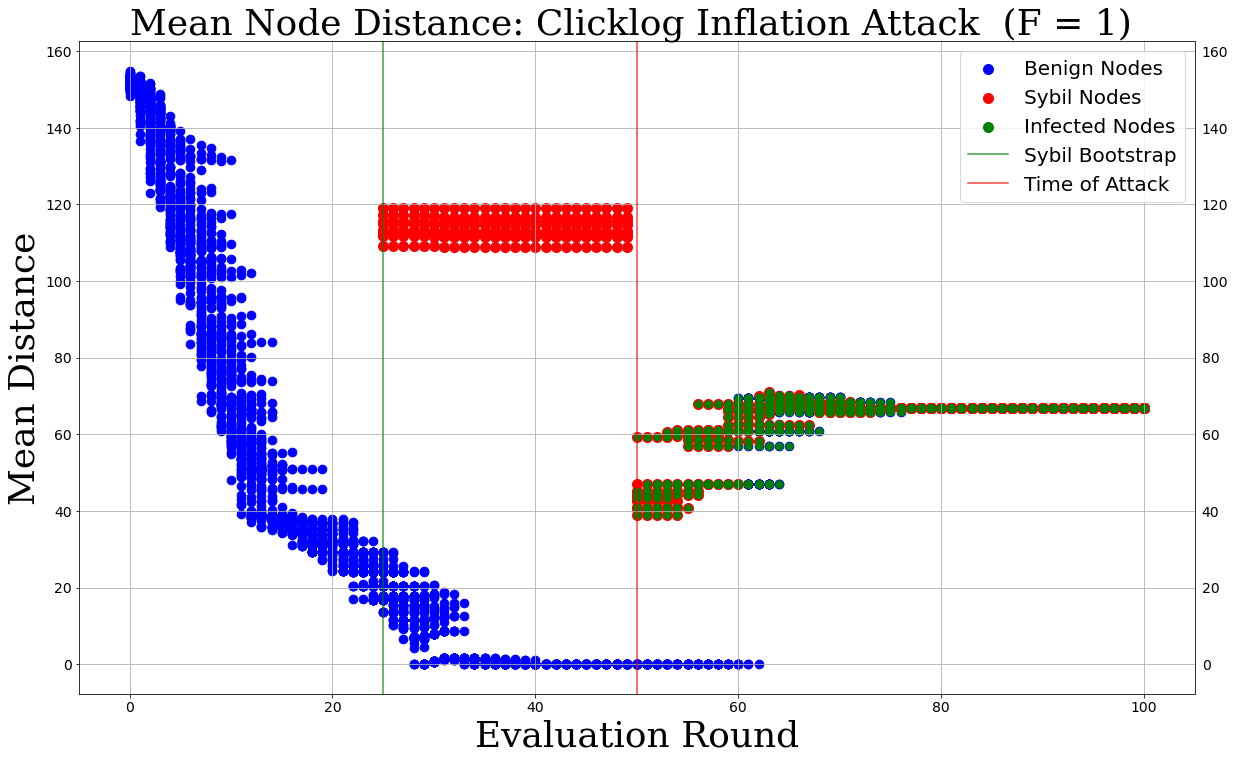}} 
\hfill
\subfloat{\includegraphics{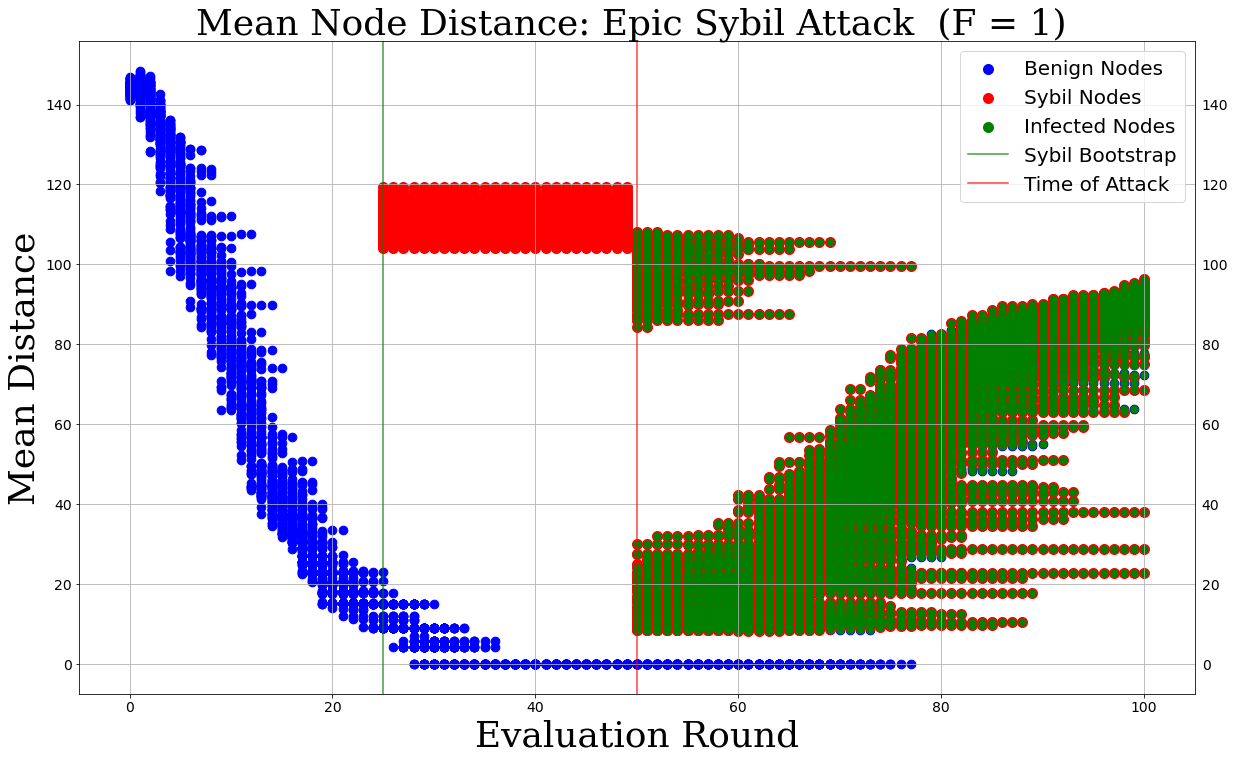}} 
\hfill

\caption{Scatter plots depicting each node's average distance to optimal rankings for each term, for each adversarial simulation, for both $F=0$ (top) and $F=1$ (bottom).}
\label{fig:main results}
\end{figure*}

\subsection{Performance Analysis}
\label{sec:analysis}

The results of the initial baseline experiment show that without any adversarial conditions the performance of G-Rank rapidly approaches the globally-optimal ranking for each node.
\hyperref[fig:baseline distance]{Figure 3} shows that the distance between each node's local ranking of results and the globally-optimal ranking for each possible query term drops precipitously in early stages of the simulation, approaching perfect ranking scores for all peers in the network. 
\hyperref[fig:median]{Figure 7} shows that the median percentage of queries containing the most popular song per tag initially grows slowly, accelerating in growth as gossip continues due to the increasing awareness of other nodes' queries and results.
As more gossip occurs, the number of queries containing the top song associated with each query approaches 100\%.
Notably, the \textit{Epic Sybil Attack} simulations were the only experiments in which the median percentage of queries did not reach 100\% by the end of the simulation, which likely is due to the significantly higher number of adversaries polluting each peer's clicklog.
\hyperref[fig:top songs]{Figure 8} shows how the number of most popular items associated with each possible query term grows at approximately even rates, indicating that the gossip scheme itself does not lead to an imbalance over time in ranking performance for lesser-used query terms.
\hyperref[fig:gossip size]{Figure 9} shows that while average gossip message size grows quickly at first, the rate of growth rapidly slows as peers become increasingly aware of one another.
The size of each gossip message rarely exceeds 600 kB.

Considering the known threat that sybil and spam attacks pose to p2p networks, the results of the adversarial simulations generally fall in line with expectations.
G-Rank is susceptible to sybil and spam attacks, though its relative resilience in the face of targeted attacks is notable.
However, both the \textit{Targeted Sybil Attack} and the \textit{Clicklog Inflation Attack} have an outsize effect on performance, where due to the sheer size of infected clicklog entries, the entire network converges towards towards a single set of rankings that it appears unable to escape from, even considering the injected noise described in \hyperref[sec:model]{Section 3A}.
This indicates that local minimums are exceedingly difficult for G-Rank to escape from.
As seen in \hyperref[fig:main results]{Figure 4}, G-Rank deviates from optimality post-attack, albeit at vastly different rates depending on the manner of adversarial interference.
\hyperref[fig:main results]{Figure 4} also shows that when $F=0$, benign users are not as quickly affected by malicious gossip, most noticeably during the \textit{Targeted Sybil Attack} and the \textit{Epic Sybil Attack}. 
Rankings in the \textit{Clicklog Inflation Attack} experience a more rapid divergence from optimality than in the \textit{Targeted Sybil Attack}, though the results of the \textit{Epic Sybil Attacks} are significantly worse.
When adversaries constitute a super-majority of peers in the network, results degrade rapidly before tapering off.
Ranking results post-attack are still significantly more accurate than at the start across all adversarial simulations.

\begin{figure}
    \centering
    \includegraphics[scale=0.2]{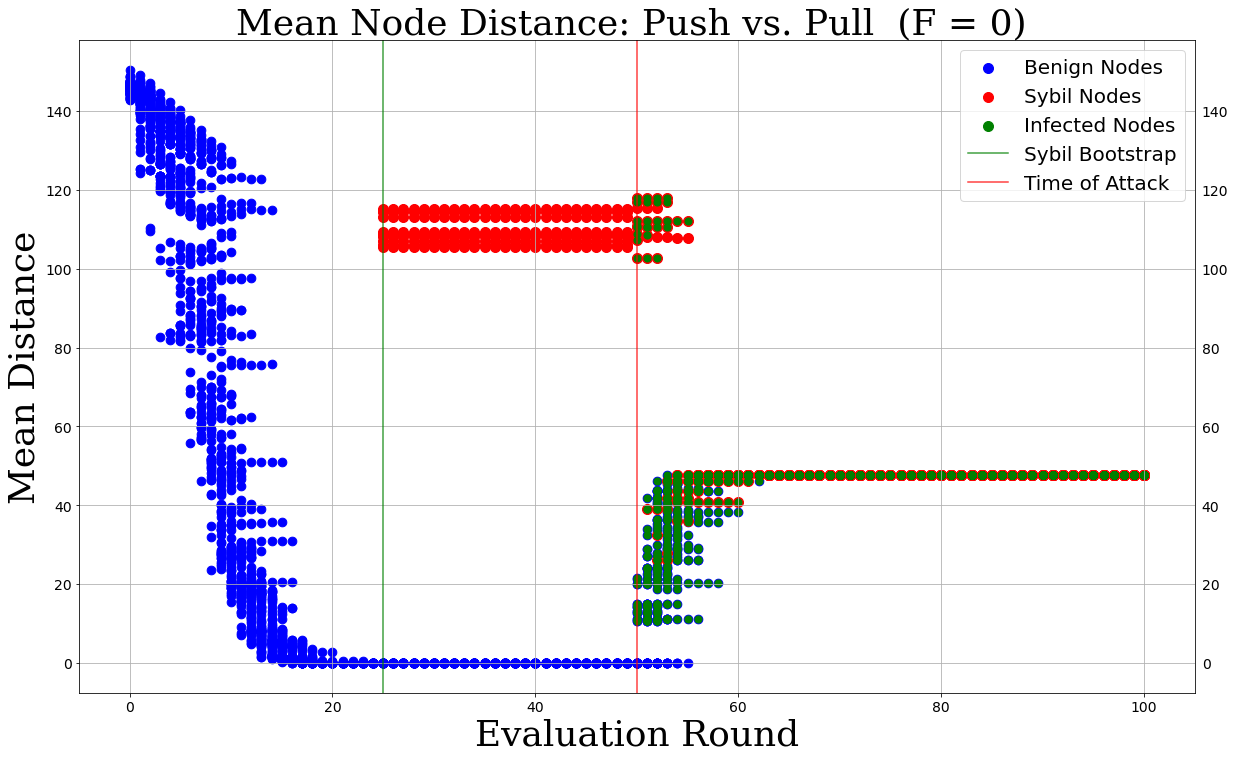}
    \caption{\textit{Push vs. Pull}: Mean node distance to optimal rankings across all terms for $F=0$.}
    \label{fig:push f0}
\end{figure}

\begin{figure}
    \centering
    \includegraphics[scale=0.2]{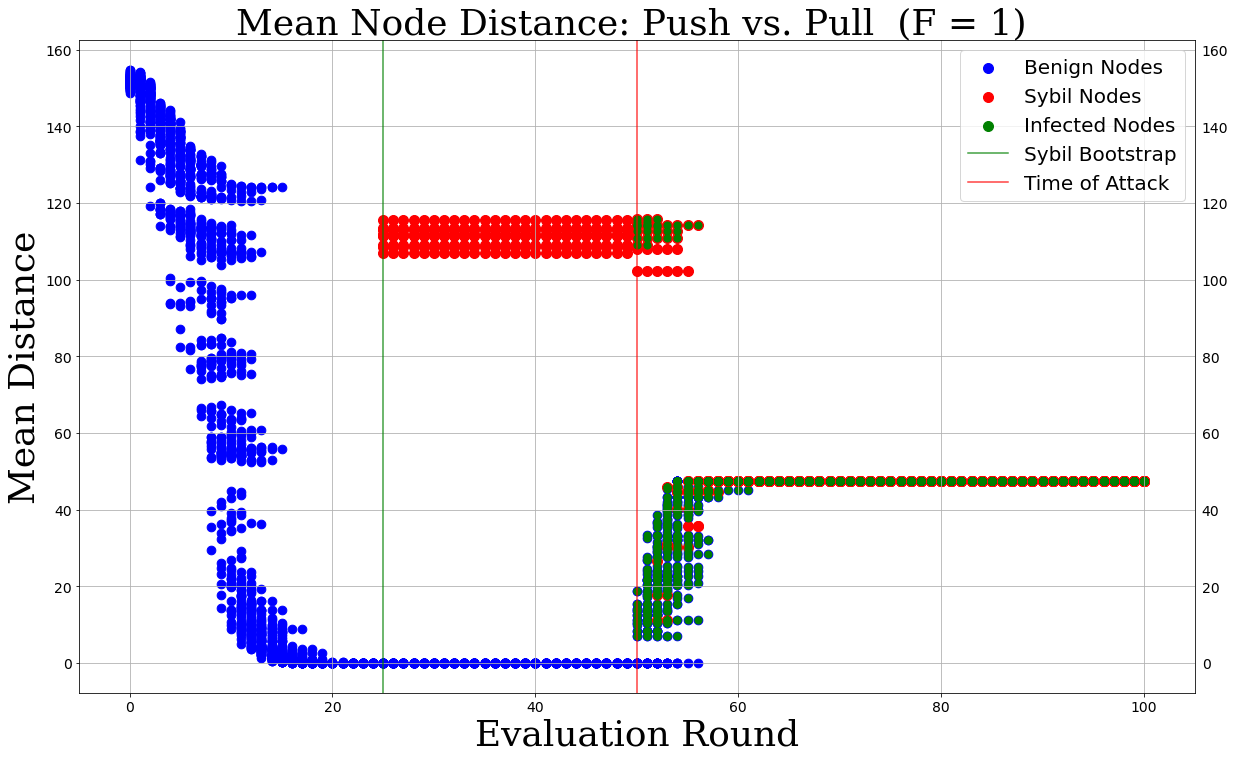}
    \caption{\textit{Push vs. Pull}: Mean node distance to optimal rankings across all terms for $F=0$.}
    \label{fig:push f1}
\end{figure}

Notably, network peers in the \textit{Push vs. Pull} simulation see ranking performance rates improve faster than in the \textit{Targeted Sybil Attack} with the pull-based gossip scheme, as seen in \hyperref[fig:push f0]{Figure 5} and \hyperref[fig:push f1]{Figure 6}. 
The \textit{Push vs. Pull} comparison demonstrates that push-based gossip schemes result in faster convergence at the expense of faster divergence under adversarial influence.
In both versions of this simulation, G-Rank converges significantly faster towards optimality, though is almost immediately trapped in a local minimum post-attack, further bolstering the argument that the gossip dissemination scheme holds an outsize influence on G-Rank's overall performance and resilience.

The effect that the isolation constant $F$ has on such behavior is minimal, but not negligible.
Setting $F=0$ has the consequence of effectively disqualifying any nodes without at least one matching query-click pair with the querying node.
Any influence an adversary then has on ranking is dependent upon the number of matching query-click pairs.
As such, as the diversity of clicklog entries grows larger so too should the effect such a parameter has on insulating nodes from other malicious peers, in theory.
Conversely, when $F>0$, all clicklog data (including malicious entries) is considered in the ranking process as the weight of each entry will subsequently also be a positive non-zero value.
The positive effect this has is greater personalization results for benign queries; clicklog results from nodes with similarity scores $S_i(n_j)=0$ still have the number of clicks associated with that result considered in the final ranking.
The negative effect is that all clicklog entries, including malicious entries, are considered.
In our simulations, there exists a negative effect on rank distances when $F>0$, although the effect is not enough to permanently isolate peers from adversarial clicklog poisoning.

\begin{figure}
    \centering
    \includegraphics[scale=0.2]{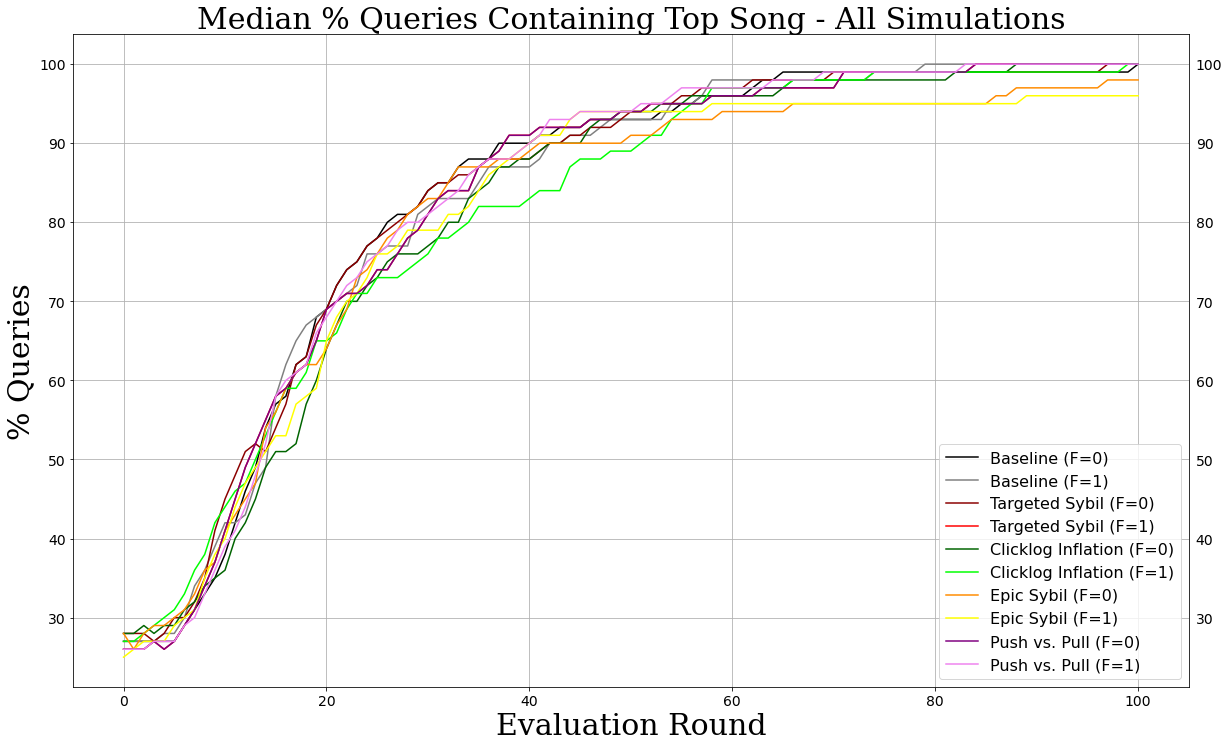}
    \caption{The median percentage of queries containing the most popular song for each query, across all experiments.}
    \label{fig:median}
\end{figure}

\begin{figure}
    \centering
    \includegraphics[scale=0.32]{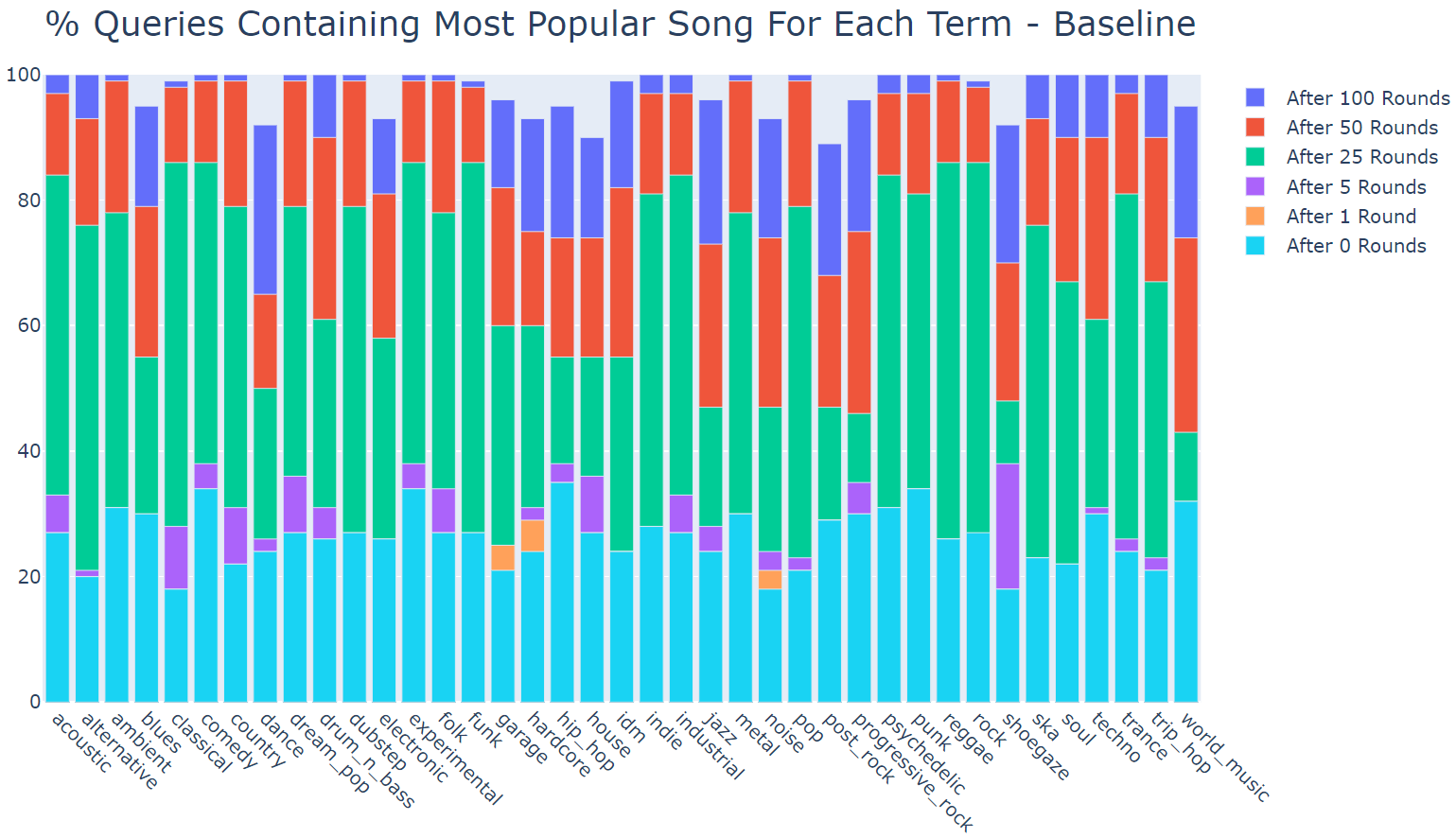}
    \caption{Percentage of queries where the most popular song associated with each query term is included in the ranked results, average between both baseline simulations.}
    \label{fig:top songs}
\end{figure}

\begin{figure}
    \centering
    \includegraphics[scale=0.2]{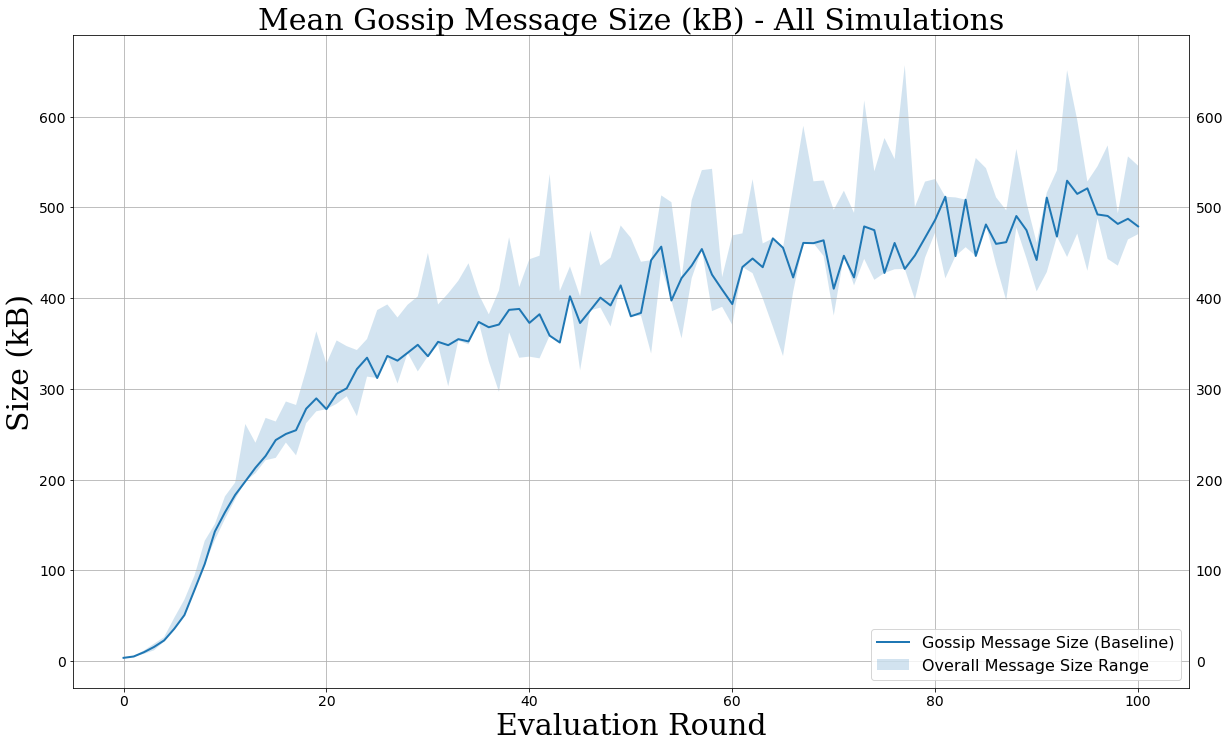}
    \caption{Mean gossip message size (in kilobytes) for all simulations. Each clicklog row entry is approx. 600 bytes in size.}
    \label{fig:gossip size}
\end{figure}


\section{Conclusion}
\label{sec:conclusion}

We proposed G-Rank as a lightweight, modular, and easy-to-understand unsupervised continuous ranking model designed explicitly for permissionless p2p networks.
The results demonstrate that unsupervised search-and-rank models designed specifically for p2p applications show merit and are worthy of further research.
G-Rank demonstrates that a simple unsupervised model can recommend near-perfect results to users in sterile network conditions.
The self-clustering method described in \hyperref[sec:similarity]{Section 3D} allows for a high degree of algorithm customization such that individual nodes can dramatically alter their ranked results based on the behavior of other nodes.
By altering their search results based on similarity of behavior, peers in the network are able to isolate themselves from adversarial behavior to varying degrees.
G-Rank shows varying degrees of resilience in the face of transient adversarial conditions, particularly regarding highly targeted nefarious behavior. 
The scale of negative adversarial impact depends heavily upon the clicklog dissemination gossip scheme.
Furthermore, the isolation constant $F$ has some effect on insulating peers from adversarial clicklog poisoning, though we suggest further research including larger more diverse data such that peers can more effectively distance themselves from the behavior of dissimilar peers.

Potential for future development of unsupervised decentralized search and ranking models in p2p networks is exceptionally rich.
One of the primary pitfalls of the p2p network domain is the threat adversarial actors such as sybil attackers may have on the model.
As such, mitigating threats above the network and protocol layers at the model level is a rich field for future development.
Other potential model improvements may include augmenting the user clustering model beyond a simple similarity metric such that sybil and other spam attacks become classified as outliers with regards to "typical" user behavior, where recommendation and ranking scores are more heavily influenced by the behavior of other users within clusters.
Fuzzy clustering methods such as the one mentioned herein allow for peers to improve their local rankings based on those received by other similar users.
More explicit self-clustering methods may lead to significantly improved performance, particularly those adept at identifying and isolating statistical outliers, such as measuring the distance of an outlier from all known cluster centroids.
We therefore conclude that unsupervised learn-to-rank models in adversarial p2p networks show significant promise and are worthy of further research.

\addtolength{\textheight}{0cm}   





\end{document}